\documentclass[10pt,twocolumn,letterpaper]{article}

\usepackage{cvpr}              

\definecolor{cvprblue}{rgb}{0.21,0.49,0.74}
\usepackage[pagebackref,breaklinks,colorlinks,allcolors=magenta]{hyperref}

\def\paperID{*****} 
\def\confName{CVPR}
\def\confYear{2025}

\title{NTIRE 2025 Challenge on Short-form UGC Video Quality Assessment and Enhancement: KwaiSR Dataset and Study}

\author{Xin Li\textsuperscript{1}$^{*,\dag}$, Xijun Wang\textsuperscript{1},  Bingchen Li\textsuperscript{1}, Kun Yuan\textsuperscript{2}$^{\dag}$, Yizhen Shao\textsuperscript{2} \\ Suhang Yao\textsuperscript{1}, Ming Sun\textsuperscript{2}, Chao Zhou\textsuperscript{2}, Radu Timofte\textsuperscript{3} and Zhibo Chen\textsuperscript{1} \\
\textsuperscript{1}University of Science and Technology of China  \quad 
\textsuperscript{2}KuaiShou Technology \quad
\textsuperscript{3}University of W\"urzburg
}

\begin{document}
\twocolumn[{
\renewcommand\twocolumn[1][]{#1}%
\maketitle
\begin{center}
\captionsetup{type=figure}
\includegraphics[width=0.96\textwidth]{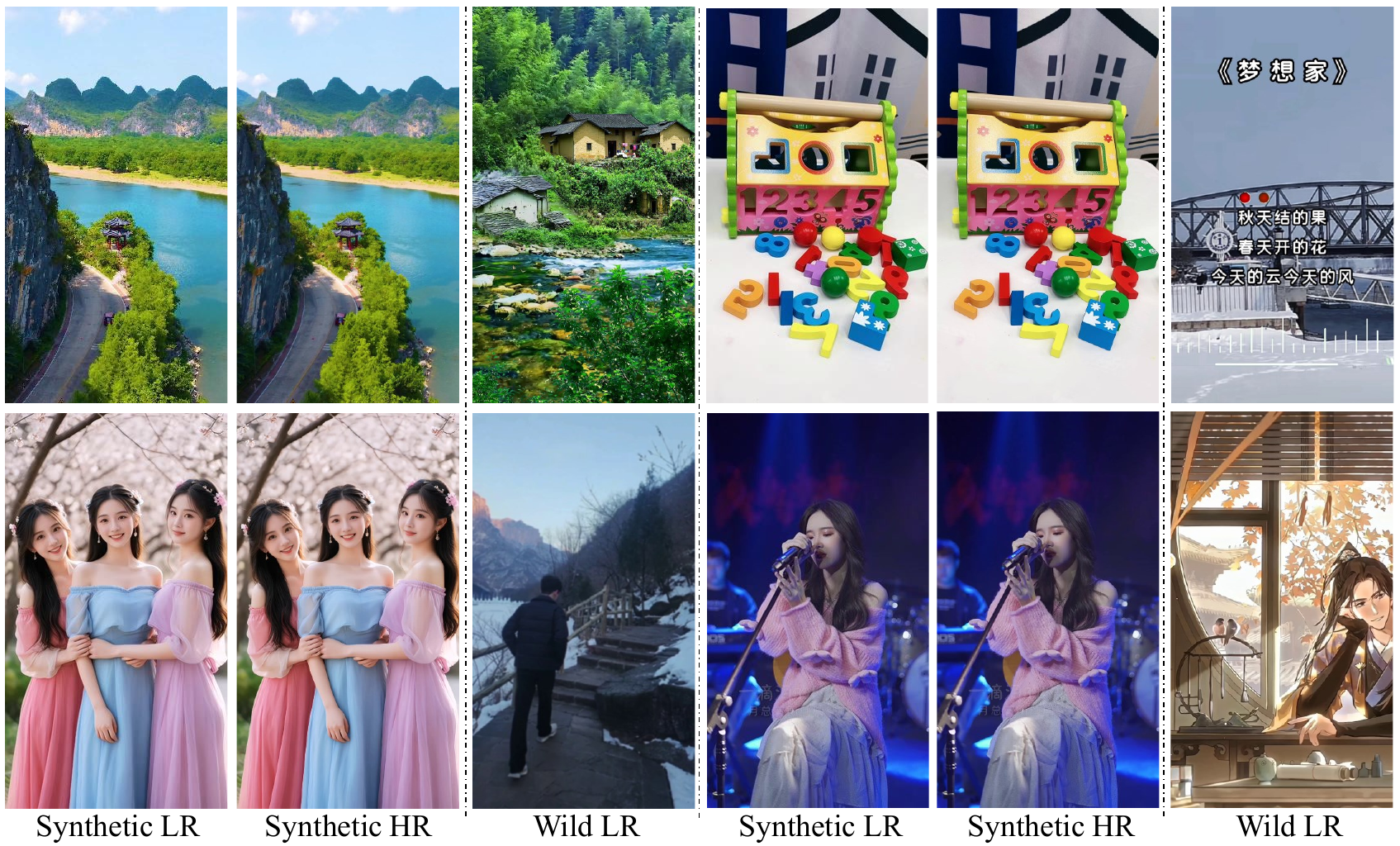}
    \captionof{figure}{The visualization of our proposed KwaiSR dataset.}
    \label{fig:demo}
\end{center}
}] 
\maketitle
\renewcommand{\thefootnote}{}
\footnotetext{$^{*}$ Corresponding author.}
\footnotetext{$^{\dag}$ X. Li and K. Yuan are the project leaders.}
\begin{abstract}
In this work, we build the first benchmark dataset for short-form UGC Image Super-resolution in the wild, termed KwaiSR, intending to advance the research on developing image super-resolution algorithms for short-form UGC platforms. This dataset is collected from the Kwai Platform, which is composed of two parts, \ie, synthetic and wild parts. Among them, the synthetic dataset, including 1,900 image pairs, is produced by simulating the degradation following the distribution of real-world low-quality short-form UGC images, aiming to provide the ground truth for training and objective comparison in the validation/testing. The wild dataset contains low-quality images collected directly from the Kwai Platform, which are filtered using the quality assessment method KVQ from the Kwai Platform. As a result, the KwaiSR dataset contains 1800 synthetic image pairs and 1900 wild images, which are divided into training, validation, and testing parts with a ratio of 8:1:1. Based on the KwaiSR dataset, we organize the NTIRE 2025 challenge on a second short-form UGC Video quality assessment and enhancement, which attracts lots of researchers to develop the algorithm for it. The results of this competition have revealed that our KwaiSR dataset is pretty challenging for existing Image SR methods, which is expected to lead to a new direction in the image super-resolution field. The dataset can be found from~\url{https://lixinustc.github.io/NTIRE2025-KVQE-KwaSR-KVQ.github.io/}.
\end{abstract}    
\section{Introduction}
\label{sec:intro}
Image Super-resolution (SR) has been a long-term research topic in low-level vision, which aims to restore the vivid textures of low-resolution images and super-resolve the resolution of images simultaneously~\cite{ATD,swinir,SR3+,HAT,li2024sed,OSEDiff,li2024ucipMLP}. Notably, the capability of image SR models is decided by two essential parts: (i) contextual/relation modeling capability between different pixels; and (ii) restoration priors learned from Image super-resolution datasets~\cite{wei2020aim}. To improve the contextual modeling probability, existing SR works are devoted to developing the frameworks based on different backbones, including Convolutional neural networks (CNNs)~\cite{SRCNN,EDSR,SR2——VDSR-cnn,RCAN,li2020learning,li2024lossagent}, Transformer~\cite{swinir,swinIR-transformer,sr_transformer1,sr2_transformer2,sr2_transformer3}, Multi-layer Perception (MLP)~\cite{tu2022maxim,li2024ucipMLP}, and Mamba~\cite{guo2024mambair,ren2024mambacsr,mambasr1,mambasr2,mambasr3,guo2024mambairv2}. Among them, CNN-based SR methods are better at capturing the local dependencies between different pixels, whereas global contextual modeling relies on enlarging the kernel size and layer depth. In contrast, the Transformer is designed based on multiple self-attention layers, which can model the dependencies between any two pixels, thereby achieving great performance on the Image SR task. However, it is resource-costly for the computation of self-attention in the Transformer. To eliminate this, MLP and Mamba-based SR methods are proposed. MLP-based methods typically achieve global contextual modeling with window-based partitions or pixel shift operations, while Mamba-based methods usually rely on scanning strategy designs and adaptive memorized mechanisms to reduce computational cost. 

 The above designs ignore the storage and re-utilization of restoration priors among datasets. With the development of generative models, some works began to explore how to utilize the generative priors existing in the GAN-based framework and diffusion-based framework to restore the vivid textures lost in low-resolution images. Since the unstable training of GAN, diffusion-based generative models have become mainstream in image/video generation, restoration, editing, etc. Early works~\cite{DiffBIR,StableSR,yu2024scaling-SUPIR,qu2024xpsr,li2023diffusion,ren2024moediff} on the diffusion-based SR method need to be retained with large-scale image super-resolution datasets. With the development of efficient transfer learning, some works take the first step to introduce the ControlNet~\cite{zhang2023adding-ControlNet} or LoRA~\cite{hu2022lora} to the pre-trained diffusion models, adapting them to downstream image restoration tasks. Furthermore, SUPIR~\cite{yu2024scaling-SUPIR} reveals the importance of the dataset for the generalization capability of diffusion-based methods in the wild. 

However, the above methods are devoted to investigating the traditional image super-resolution task. With the development of short-form UGC platforms~\cite{lu2024kvq,li2024ntireKVQ-Challengereport,blau20182018PIRMchallenge}, like Kwai and TikTok, the restoration of short-form UGC images/videos has become an emergent task for improving the user experience since most images/videos are captured and uploaded by unprofessional/inexperienced users. However, there are almost no professional datasets that have been collected to evaluate the effectiveness of the restoration of existing algorithms on short-form UGC image restoration. To overcome this, we build the first benchmark dataset, \ie, the KwaiSR dataset, for the short-form UGC image super-resolution task, and organize the challenge on diffusion-based image super-resolution together with the NTIRE workshop~\cite{ntire2025shortugckvq-challengereport}.

\begin{figure*}
    \centering
    \includegraphics[width=0.85\linewidth]{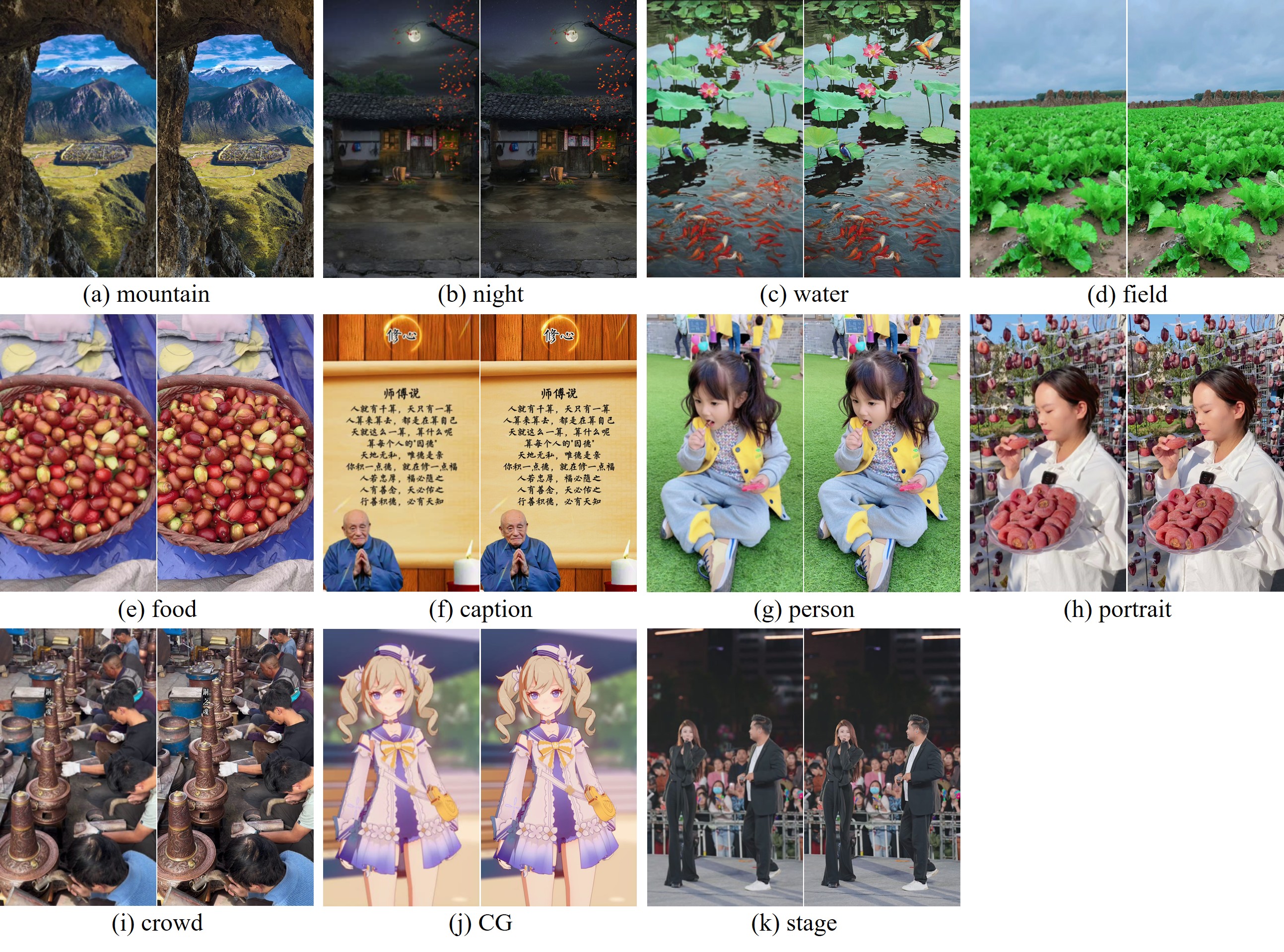}
    \vspace{-2mm}
    \caption{The eleven semantic categories of the KwaiSR dataset. The left/right images are LR/HR, respectively.}
    \label{fig:sementics}
\end{figure*}

Concretely, in contrast to existing SR datasets, which are composed of either synthetic image pairs for image SR or real-world images without ground truth. Our KwaiSR dataset is composed of two parts, \ie, the synthetic part and the wild part. The synthetic part is simulated to satisfy the distribution of the degradation existing in real-world short-from UGC images, resulting in 1800 image pairs. Since it is composed of low-resolution images and corresponding high-resolution images, researchers can utilize it for the development of the SR methods for short-form UGC image super-resolution. For the wild part, we directly utilize the quality assessment method KVQ from the Kwai Platform to select the low-quality images in the wild, resulting in 1900 images for development and validation. To keep the diversity of semantics, we select the source images from 11 classes, including mountain, night, water, field, food, caption, person, portrait, crowd, CG, and stage, which can be observed from Fig.~\ref{fig:sementics}. The degradation degree of images in this dataset is also considered to keep balanced when collecting the datasets. 

The KwaiSR dataset is randomly divided into training, validation, and testing parts with a ratio of 8:1:1. To systematically investigate the characteristics of our proposed KwaiSR dataset, we analyze the distribution of degradation degree and validate the effectiveness of existing diffusion-based methods on this dataset. Moreover, from the competition, we can find that this dataset is pretty challenging for recent techniques to improve subjective quality. The conclusion is as: (i) The trade-off between realism and perceptual quality is challenging in this dataset; (ii) Existing metrics are not accurate to measure the human perception quality in this dataset; (iii) Existing diffusion-based methods are not effective enough for this dataset. 
\section{Related Work}
\label{sec:related work}

\subsection{Image Super-Resolution Dataset}

\noindent \textbf{The training datasets} are fundamental to the success of image super-resolution (ISR) tasks. DIV2K~\cite{DIV2K} proposes a high-quality image dataset with a maximum side resolution exceeding 2k, with 800 images for training, 100 images for validation, and another 100 images for testing. Notably, the 100 testing images are only accessible as downsampled low-resolution images, with their high-resolution counterparts remaining unavailable. To further boost the performance of ISR models, Timofte~\etal~\cite{Flickr2K} collect 2650 high-resolution images from the Flickr website with quality on par with DIV2K, obtaining the Flickr2K~\cite{Flickr2K} dataset. To extend the diversity of outdoor scenes, Wang~\etal propose the OST~\cite{SFTGAN_OST} dataset with seven outdoor scene categories, including water, sky, plant, mountain, grass, building, and animal. More recently, Li~\etal collected a larger dataset LSDIR~\cite{li2023lsdir} which contains 84991 images for training, 1000 images for validation, and 1000 images for testing. Yang~\etal introduce HQ-50K~\cite{hq-50k} with 50k high-quality images from various semantic categories for training. 

However, these datasets lack distortion pairs captured in \textbf{real-world} scenarios, making them unsuitable for training super-resolution models for real-world applications. RealSR~\cite{RealSR} and DRealSR~\cite{DrealSR} are two typical super-resolution datasets captured in real-world scenarios using DSLR cameras. The authors of these datasets captured LR-HR pairs by adjusting the camera's focal length and then generated super-resolution image pairs for training and testing through careful alignment algorithms. 

\noindent \textbf{The evaluation benchmarks} are essential for assessing the performance of ISR methods. While some of the aforementioned datasets provide corresponding validation splits, they still fail to comprehensively validate the performance of ISR methods. For classical ISR problems that typically consider Bicubic downsampling distortions, there are five commonly used benchmarks, such as Set5~\cite{Set5}, Set14~\cite{Set14}, BSD100~\cite{BSDall}, Urban100~\cite{urban100}, and Manga109~\cite{manga109}. For real-world ISR problems, the popular evaluation benchmarks are RealSR and DRealSR, as they are captured naturally in real-world scenarios. Apart from these two datasets, there are several benchmarks with only LR images available, such as OST300~\cite{SFTGAN_OST} and RealSRSet~\cite{BSRGAN}. The no-reference (NR) image quality assessment (IQA) metrics~\cite{qinstruct,yu2024sf,lu2024aigc,co-instruct,deqa,wu2023qalign,depictqa} are widely chosen to evaluate performance on benchmarks without HR.

\subsection{Image Super-Resolution Backbones}

Before the evolution of deep neural network (DNN)-based algorithms, traditional numerical methods were often used to tackle image super-resolution problems, such as Bilinear, Bicubic, and Lanczos upsampling algorithms. With the development of convolutional neural networks (CNN), SRCNN~\cite{SRCNN}, as a pioneering work, introduces CNN to tackle ISR problems by minimizing the mean squared error (MSE) between the super-resolved images (SR) and their corresponding HR counterparts. To further enhance the network's representation capability, researchers adopt residual connections~\cite{EDSR}, dense connections~\cite{RDN}, and multi-scale representations~\cite{MSRN,MSDNN}, \etc. Additionally, some methods~\cite{RCAN,SAN} leverage attention mechanisms to improve the reconstruction of high-frequency textures. 

With the advent of vision transformers (ViTs), CAT~\cite{CAT} adopts a cross aggregation mechanism to address ISR problems. SwinIR~\cite{swinir} incorporates powerful shifted window-based attention to better super-resolve LR images. HST~\cite{HST} develops a hierarchical attention structure for compressed ISR problems. HAT~\cite{HAT} enhances the attention mechanism by introducing overlapped windows. Compared to CNN-based methods, transformer-based methods offer superior long-range dependency modeling capabilities. By combining transformer and CNN-based modules, such as channel attention, the performance of models~\cite{ATD,ART} can be further improved.

\begin{figure*}
    \centering
    \includegraphics[width=1\linewidth]{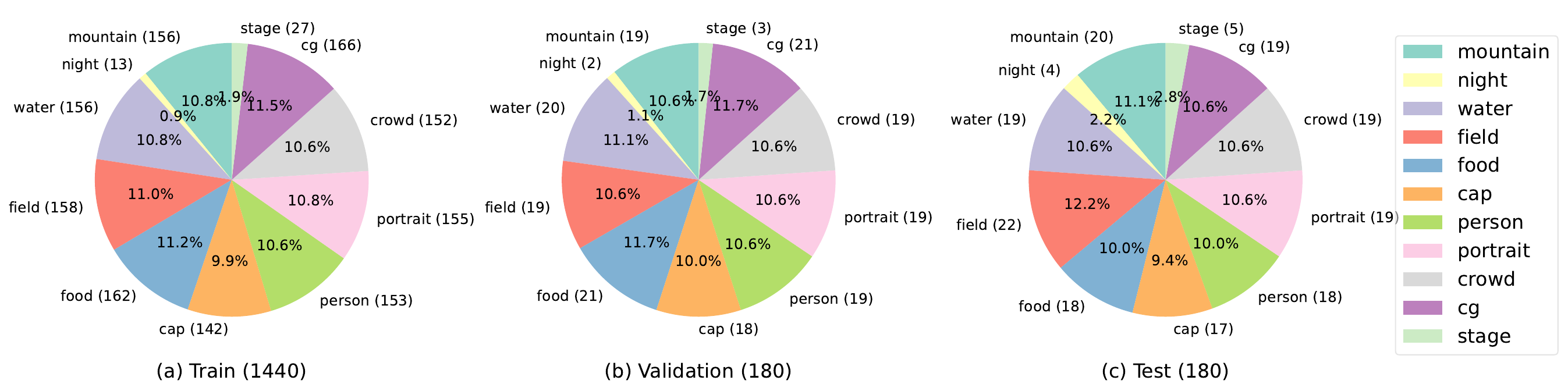}
    \caption{The semantic categories and their corresponding image counts for synthetic datasets.}
    \label{fig:syntheticpie}
\end{figure*}

\begin{figure*}
    \centering
    \includegraphics[width=1\linewidth]{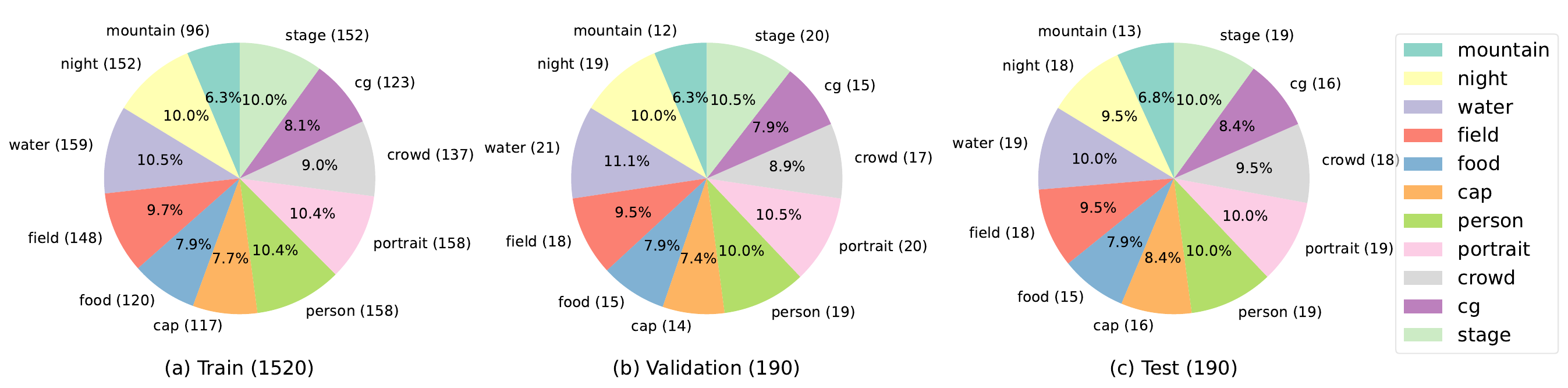}
    \caption{The semantic categories and their corresponding image counts for wild datasets.}
    \label{fig:wildpie}
\end{figure*}
\subsection{Diffusion-based Image Super-resolution}
Recently, generative methods have emerged as a promising approach for image super-resolution, leveraging the powerful generative capabilities of denoising diffusion probabilistic models~\cite{DDPM,li2023diffusion}.
These methods, such as StableSR~\cite{StableSR}, DiffIR~\cite{DiffIR}, and DiffBIR~\cite{DiffBIR}, have achieved remarkable performance in terms of perceptual quality, utilizing powerful text-to-image pre-trained models like Stable Diffusion (SD).
Specifically, StableSR introduces a time-aware encoder for low-quality (LQ) image conditioning and employs feature warping to balance fidelity and perceptual quality. PASD~\cite{PASD} leverages high-level semantic information to guide the super-resolution process, injecting information through a pixel-aware cross-attention mechanism.
Despite the significant progress in visual quality achieved by current methods, they still have some limitations, such as the need for multiple sampling steps and heavy dependence on substantial computational resources.
Recently, OSEDiff~\cite{OSEDiff} proposed a one-step sampling diffusion model, which uses variational score distillation in the latent space to conduct KL-divergence regularization. This approach reduces the number of denoising steps while maintaining high-quality visual effects.
AdcSR~\cite{AdcSR} further distills the one-step diffusion network OSEDiff into a streamlined diffusion-GAN model, pruning redundant modules to reduce model parameters. This results in a lightweight super-resolution model while retaining the visual quality of diffusion models.

\section{Dataset Analysis}
\subsection{Content Analysis}
We present the distribution of the semantics in our dataset in Fig.~\ref{fig:syntheticpie} and Fig.~\ref{fig:wildpie}. During the dataset collection, we unified the number of images that belong to different scenarios, as demonstrated in the distribution of the wild dataset. However, in the selection of high-quality paired images for the synthetic dataset, the number of images in the ``night'' and ``stage'' categories is smaller compared to other scenarios, as paired images in these categories are typically resource-intensive to collect and often yield a relatively low KVQ score. In the collection of datasets in the wild scenarios, the semantic distribution tends to be more balanced since only the distorted images need to be collected. Additionally, the training, validation, and testing datasets all share the same semantic distribution.

\begin{table*}[h]
    \centering
    \caption{Quantitative comparison with state-of-the-art methods on synthetic test set.}
    \resizebox{0.90\textwidth}{!}{\begin{tabular}{c|ccccccccc}
        \toprule
        \textbf{Metrics} & HAT~\cite{HAT} & SwinIR~\cite{swinir} & ResShift~\cite{Resshift} & StableSR~\cite{StableSR} & DiffIR~\cite{DiffIR} & DiffBIR~\cite{DiffBIR} & PASD~\cite{PASD} & OSEDiff~\cite{OSEDiff} & InvSR~\cite{InvSR}\\
        \midrule
        PSNR $\uparrow$ &27.3436 & 27.261 & 27.1563 & 27.3486 & \textbf{27.9864} & 27.4830 & 27.2658 & 25.6713 & 25.0871\\
        SSIM $\uparrow$ &0.77620 & 0.78664 & 0.77560 & 0.79463 & \textbf{0.79653} & 0.76572 & 0.77403 & 0.76090 & 0.74759\\
        LPIPS $\downarrow$ &0.35410 & 0.23076 & 0.25684 & 0.22608 & \textbf{0.22103} & 0.24288 & 0.23215 & 0.22353 & 0.23448\\
        MUSIQ $\uparrow$ &45.3917 & 65.9748 & 66.0079 & 62.1635 & 66.7515 & 72.3388 & 69.7014 & \textbf{72.7883} & 73.0559\\
        CLIPIQA $\uparrow$ &0.41546 & 0.56242 & 0.66599 & 0.55735 & 0.55046 & 0.74526 & 0.62799 & 0.70243 & \textbf{0.72408} \\
        MANIQA $\uparrow$ &0.31151 & 0.38265 & 0.39400 & 0.34637 & 0.37255 & \textbf{0.52215} & 0.42347 & 0.46209 & 0.46011\\
        \bottomrule
    \end{tabular}}
    \label{tab:comparison1}
\end{table*}

\begin{table*}[h]
    \centering
    \caption{Quantitative comparison with state-of-the-art methods on synthetic val set.}
    \resizebox{0.95\textwidth}{!}{\begin{tabular}{c|ccccccccc}
        \toprule
        \textbf{Metrics} & HAT~\cite{HAT} & SwinIR~\cite{swinir} & ResShift~\cite{Resshift} & StableSR~\cite{StableSR} & DiffIR~\cite{DiffIR} & DiffBIR~\cite{DiffBIR} & PASD~\cite{PASD} & OSEDiff~\cite{OSEDiff} & InvSR~\cite{InvSR} \\
        \midrule
        PSNR $\uparrow$ & 26.9509 & 27.0131 & 26.9797 & 27.0502 & \textbf{27.7111} & 27.2332 & 27.0005 & 25.4596 & 24.8241\\
        SSIM $\uparrow$ & 0.77251 & 0.78649 & 0.77649 & 0.79377 & \textbf{0.79508} & 0.76431 & 0.77260 & 0.76151 & 0.74608\\
        LPIPS $\downarrow$ & 0.35130 & 0.22435 & 0.25441 & 0.22349 & \textbf{0.22079} & 0.24435 & 0.22965 & 0.22371 & 0.23377\\
        MUSIQ $\uparrow$ & 45.8019 & 66.5914 & 66.4147 & 62.6499 & 67.1753 & 72.3233 & 69.6989 & 72.7677 & \textbf{73.2260} \\
        CLIPIQA $\uparrow$ & 0.42147 & 0.56816 & 0.66677 & 0.55443 & 0.54999 & \textbf{0.74392} & 0.62408 & 0.70467 & 0.72863\\
        MANIQA $\uparrow$ & 0.31923 & 0.38757 & 0.39372 & 0.34971 & 0.37195 & \textbf{0.52063} & 0.42303 & 0.46379 & 0.46554\\
        \bottomrule
    \end{tabular}}
    \label{tab:comparison2}
\end{table*}

\begin{figure}
    \centering
    \includegraphics[width=0.95\linewidth]{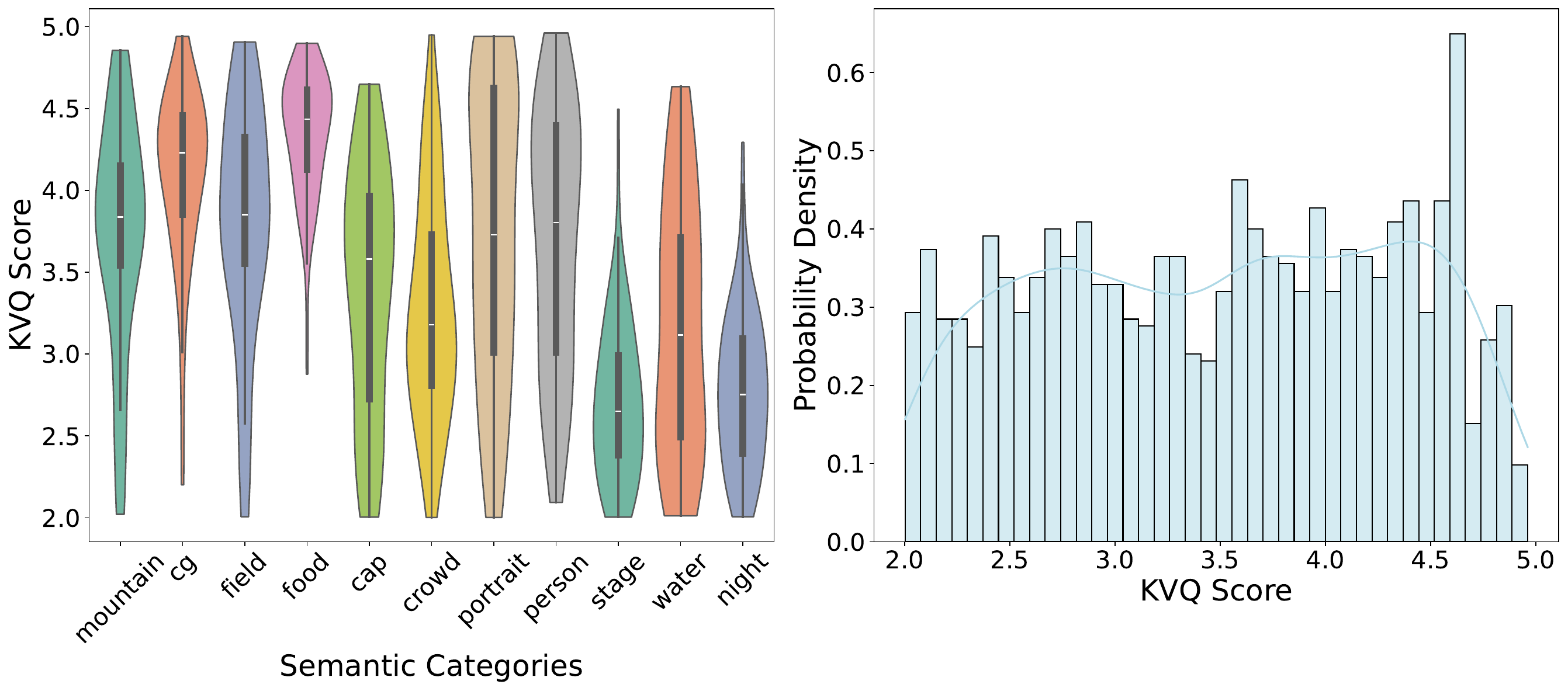}
    \caption{The KVQ Score distribution of different semantic categories
(left) and the histogram of the overall KVQ Score distribution (right) of the wild train dataset.}
    \label{fig:wild_train}
\end{figure}

\subsection{Quality Analysis}
We conduct an analysis of the quality distribution for the wild part of our KwaiSR dataset. As shown in Fig.~\ref{fig:wild_train}, it is evident that the perception scores of low-quality images are fairly consistent across the range of $[2.0-5.0]$. While the score range varies slightly across different semantics, the overall distribution of images aligns with that observed on the Kwai Platform. Notably, images in night and stage scenarios tend to have a narrower quality score range compared to other scenarios. Images with the ``food'' semantic generally exhibit higher quality in the wild setting. Based on this analysis, we conclude that this dataset is both practical and highly relevant for real-world short-form UGC platforms, making it an ideal benchmark for evaluating algorithms.

\subsection{Challenge Analysis}
This dataset is utilized for the NTIRE 2025 challenge on Short-form UGC Video Quality Assessment and Enhancement - Track 2. In this track, we aim to improve the subjective quality of low-quality short-form UGC images through diffusion-based/generation-based methods. We have received the results from 9 teams, and have conducted a user study for the top six teams based on the ranking of objective results with a composition of PSNR, SSIM, LPIPS, MUSIQ, ManIQA, and CLIPIQA metrics. We reveal that some teams excel at improving realism and fidelity, while others focus on enhancing subjective quality. More importantly, the teams optimizing the realism typically suffer from the deformation in the face, while the teams pursuing the perception quality typically generate unreal, AIGC-style textures. Another key observation is the failure of objective metrics. The top-ranked team in terms of objective scores did not deliver the best user experience, highlighting the need for a more effective quality assessment method for short-form UGC image super-resolution. Further details can be found in the corresponding challenge report~\cite{ntire2025shortugckvq-challengereport}.
\section{Experiment}
\subsection{Experimental Settings}

The proposed KwaiSR dataset is a comprehensive benchmark that includes 360 synthetic image pairs and 380 low-quality wild images. Specifically, the synthetic subset features ground-truth images with a resolution of 1920×1080, while the corresponding low-quality images are downsampled to 480×270. In contrast, the wild subset consists of low-quality images with a resolution of 1920×1080. Both subsets have been randomly split into testing and validation sets in a 1:1 ratio. During evaluation, the synthetic subset requires a 4× upsampling to match the resolution of the ground-truth images, whereas the wild subset does not require upsampling.

For reference-based metrics, we utilize three widely recognized measures: PSNR (Peak Signal-to-Noise Ratio), SSIM (Structural Similarity Index)~\cite{psnrssim}, and LPIPS (Learned Perceptual Image Patch Similarity)~\cite{lpips}. PSNR and SSIM are computed on the luminance channel in the YCbCr color space to assess the fidelity and structural similarity between the enhanced and ground-truth images. Additionally, we incorporate three state-of-the-art no-reference metrics: MUSIQ~\cite{MUSIQ}, CLIPIQA~\cite{CLIPIQA}, and MANIQA~\cite{MANIQA}, which focus on evaluating the perceptual quality of enhanced images without relying on reference images.

\subsection{Experimental Results}

For a comprehensive evaluation, we compare a variety of state-of-the-art approaches.
Specifically, we include transformer-based methods, HAT~\cite{HAT}, which combines channel attention and window-based self-attention to enhance information utilization, and SwinIR ~\cite{swinir}, which based on the Swin Transformer architecture, outperforms state-of-the-art methods in multiple tasks.

We also included diffusion-based methods, ResShift~\cite{Resshift}, StableSR~\cite{StableSR}, DiffIR~\cite{DiffIR}, DiffBIR~\cite{DiffBIR}, PASD~\cite{PASD}, OSEDiff~\cite{OSEDiff}, and InvSR~\cite{InvSR},
which have recently gained significant attention for their excellent subjective quality. 
ResShift constructed a Markov chain to transfer between high-resolution and low-resolution images through residual shifting.
StableSR designed a time-aware encoding for LQ image condition and CFW module to balance the fidelity and visual quality.
DiffBIR utilized several pre-trained cleaning modules for degradation removal and then used the diffusion model for initial generation.
PASD incorporated high-level information into the denoising process, significantly improving the perceptual quality.
And OSEDiff and InvSR reduce the sampling steps to one step through variational score distillation and noisy latent predictor, respectively.

For all these methods, we conducted tests using the official code provided and for diffusion-based methods, we set the sampling steps to the values specified in the original papers.

\begin{figure*}[htb] 
    \centering 
    \includegraphics[width=0.90\textwidth]{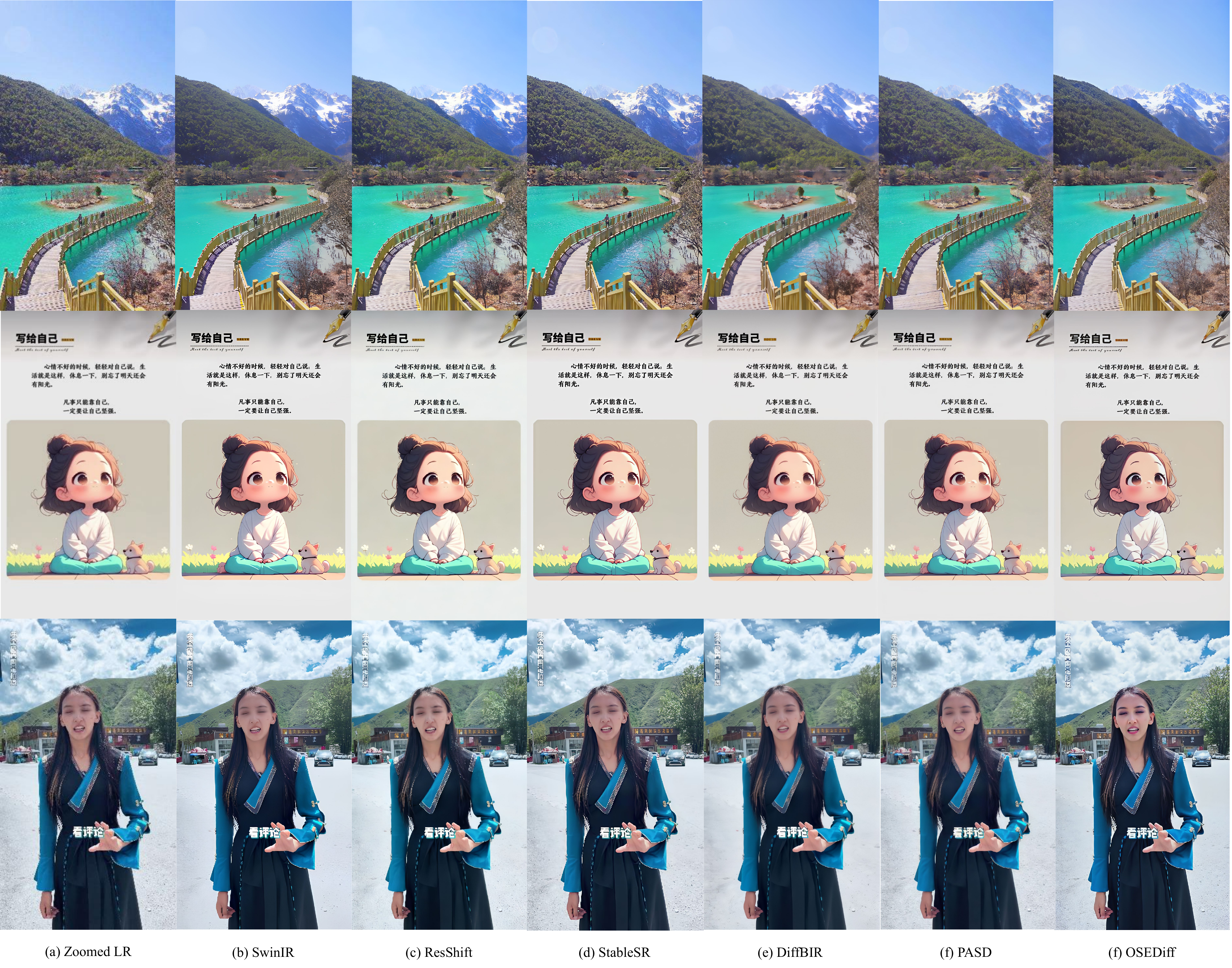} 
    \caption{Qualitative comparison with different methods. These three rows of images are sourced from the synthetic test set, synthetic validation set, and synthetic test set respectively.} 
    \label{fig:synthtic} 
\end{figure*}

\begin{figure*}[h] 
    \centering 
    \includegraphics[width=0.93\textwidth]{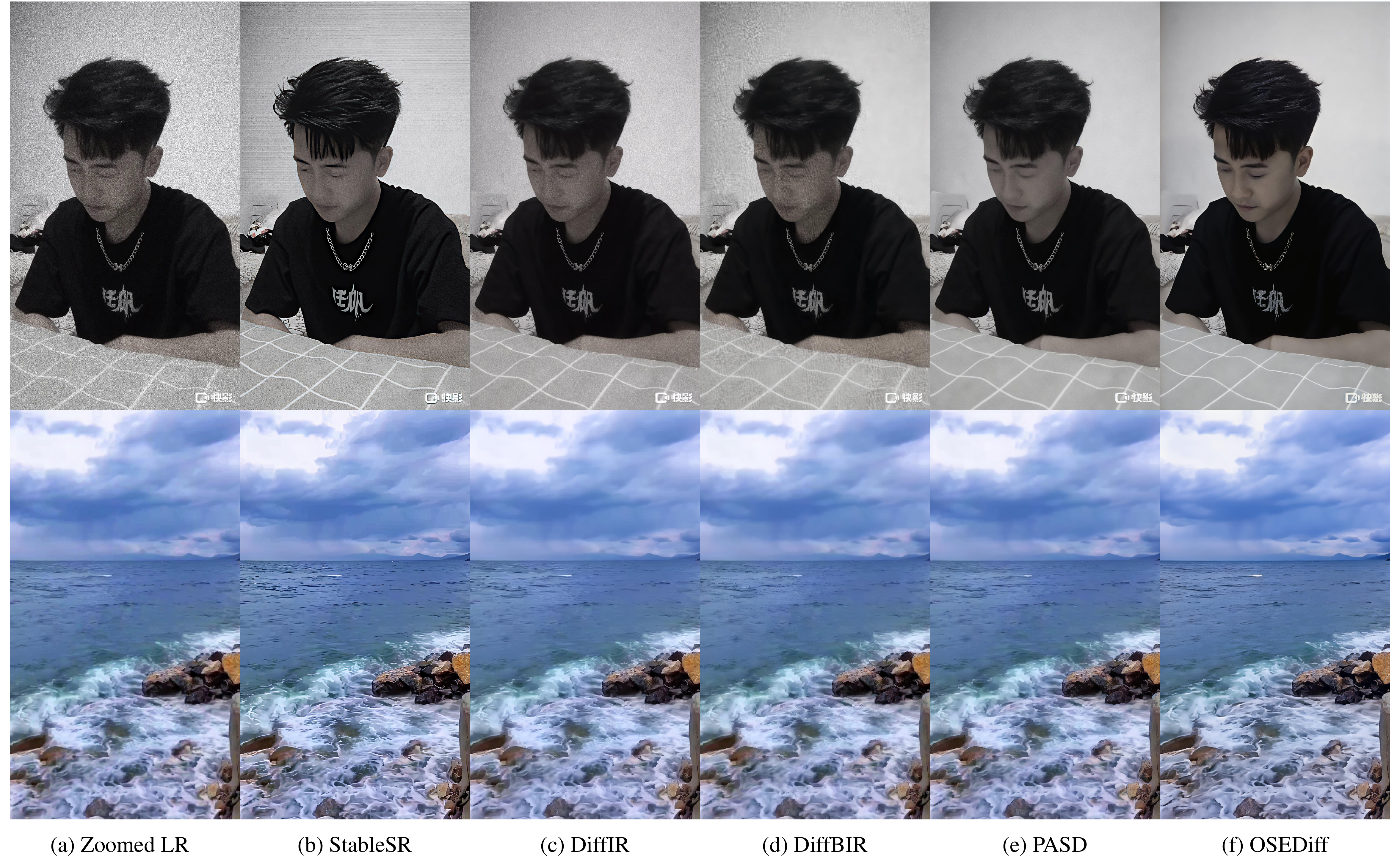} 
    \vspace{-1mm}
    \caption{Qualitative comparison with different methods. These two rows of images are sourced from the wild test set and wild validation set, respectively.} 
    \label{fig:wild} 
\end{figure*}
\begin{table*}[htp]
    \centering
    \caption{Quantitative comparison with state-of-the-art methods on wild test set.}
    \resizebox{0.88\textwidth}{!}{\setlength{\tabcolsep}{7mm}{\begin{tabular}{c|ccccc}
        \toprule
        \textbf{Metrics} & StableSR~\cite{StableSR} & DiffIR~\cite{DiffIR} & DiffBIR~\cite{DiffBIR} & PASD~\cite{PASD} & OSEDiff~\cite{OSEDiff} \\
        \midrule
        MUSIQ $\uparrow$ & 62.9131 & 57.9223 & 62.0066 & 63.5395 & \textbf{68.4065} \\
        CLIPIQA $\uparrow$ & 0.58172 & 0.44774 & 0.65425 & 0.58340 & \textbf{0.68058} \\
        MANIQA $\uparrow$ & 0.39049 & 0.33762 & \textbf{0.45201} & 0.41118 & 0.44619\\
        \bottomrule
    \end{tabular}}}
    \label{tab:comparison3}
\end{table*}

\begin{table*}[h]
    \centering
    \vspace{-3mm}
    \caption{Quantitative comparison with state-of-the-art methods on wild validation set.}
    \resizebox{0.9\textwidth}{!}{\setlength{\tabcolsep}{7mm}{\begin{tabular}{c|ccccc}
        \toprule
        \textbf{Metrics} & StableSR~\cite{StableSR} & DiffIR~\cite{DiffIR} & DiffBIR~\cite{DiffBIR} & PASD~\cite{PASD} & OSEDiff~\cite{OSEDiff} \\
        \midrule
        MUSIQ $\uparrow$ & 63.9250 & 60.4220 & 63.1168 & 65.7276 & \textbf{68.7709}  \\
        CLIPIQA $\uparrow$ & 0.60363 & 0.46825 & 0.66702 & 0.61496 & \textbf{0.69633}   \\
        MANIQA $\uparrow$ & 0.38902 & 0.34284 & \textbf{0.45658} & 0.41904 & 0.44622  \\
        \bottomrule
    \end{tabular}}}
    \label{tab:comparison4}
    \vspace{-3mm}
\end{table*}

\noindent\textbf{Quantitative Comparison.} Experiments were conducted on both the synthetic and wild subsets of the KwaiSR dataset. 
The following four tables present the quantitative results obtained on these subsets.

As shown in Table~\ref{tab:comparison1} and Table~\ref{tab:comparison2}, diffusion-based methods generally outperform transformer-based methods on the synthetic test and validation sets. 
Specifically, on the synthetic benchmark, DiffIR achieves the highest PSNR and SSIM scores, indicating its superior reconstruction quality. 
Furthermore, DiffIR also achieves the lowest LPIPS score, 
further highlighting its effectiveness in generating images that are structurally similar to the ground truth.
For perceptual quality assessment, DiffBIR, OSEDiff, and InvSR demonstrate the best overall performance across the three perceptual metrics. 
DiffBIR achieves the highest CLIPIQA and MANIQA scores, while OSEDiff attains the highest MUSIQ score on both synthetic subsets. 
However, OSEDiff's relatively lower PSNR and SSIM scores suggest that its reconstruction quality is not as strong as that of DiffIR. 
We attribute this to the fact that OSEDiff is a one-step sampling diffusion model, 
which may inherently limit its reconstruction capabilities compared to multi-step diffusion models.
InvSR, which also supports one-step sampling, exhibits the poorest performance in terms of PSNR and SSIM on both the test and validation sets. However, it achieves excellent perceptual quality. This observation suggests that our hypothesis regarding single-step sampling is valid: while fewer sampling steps can reduce reconstruction quality in terms of traditional metrics like PSNR and SSIM.

On the wild dataset, which consists of real-world images that do not require 4× upsampling, 
we evaluated only the methods that support 1x image restoration, namely StableSR, DiffIR, DiffBIR, PASD, and OSEDiff. 
As shown in  Table~\ref{tab:comparison3} and Table~\ref{tab:comparison4}, the perceptual quality of most methods has declined to some extent, especially for DiffIR and DiffBIR. 
This may be attributed to their relatively weaker generalization capabilities, 
as the wild dataset presents more diverse and complex real-world scenarios that differ from the synthetic training data.
In contrast, OSEDiff continues to demonstrate excellent performance in terms of perceptual quality. 
It maintains strong generative capabilities while employing a single-step sampling approach. 
This suggests that OSEDiff is more robust in handling real-world variations and 
achieving high perceptual quality even without the need for multiple sampling steps.



\noindent\textbf{Qualitative Comparison.}
To further evaluate the performance of various methods on the KwaiSR dataset, we present visual comparisons of different methods on the synthetic and wild subsets in Figures 1 and 2. The two rows in Figure 1 are derived from the synthetic test set and validation set, respectively.

As illustrated in the first row of Figure~\ref{fig:synthtic}, SwinIR generally produces satisfactory visual quality, but it suffers from blurring in the middle of the trees due to the lack of strong generative priors. DiffBIR and OSEDiff generate relatively smooth visual effects, failing to preserve the details on the sea surface. In contrast, PASD and ResShift retain more details and achieve the best visual quality, but their performance in the portrait area is not satisfactory. The results of StableSR still contain some noise, leading to an overall visual quality that is not as good as the other methods.

In the second row of Figure~\ref{fig:synthtic}, text processing is the main focus. PASD's text exhibits excessive abnormal structures, resulting in the worst visual appearance. Surprisingly, SwinIR, which lacks generative priors, performs better in text processing, aligning more closely with the low-resolution (LR) image.
The other methods have their strengths and weaknesses, with similar visual effects in general. It is worth noting that the results generated by Resshift show significant color shifts compared to the LR image.
For the cartoon characters in the lower part of the image, all methods have generated satisfactory visual quality.

In the third row, the various methods exhibit results similar to those in the first two rows. Notably, OSEDiff introduces additional details in the generation of the eyes that are inconsistent with the low-resolution (LR) image. This aligns with the lower PSNR and SSIM scores observed for OSEDiff in Table~\ref{tab:comparison1} and Table~\ref{tab:comparison2}.

In Figure~\ref{fig:wild}, we present visual comparisons on the wild dataset. In the first row, the LR (Low-Resolution) images contain a significant amount of noise. PASD and OSEDiff perform well in handling this noise, generating smoother results without noise. Other methods retain some of the noise from the original images to varying degrees, resulting in less-than-ideal visual quality.
Specifically, StableSR remains some noise on the face and hand and has some color shifting on the cloth icon and necklace.
In the second row, where the LR images have relatively good quality, the visual performance of all methods is comparatively better. They are all capable of effectively restoring details in the waves and the horizon, thereby enhancing the overall image quality.
Overall, the KwaiSR dataset highlights the complexities of real-world super-resolution tasks, with challenges including detail preservation, noise handling, and color accuracy.

\section{Conclusion}
In this work, we introduce KwaiSR, the first benchmark dataset for short-form UGC Image Super-Resolution in the wild. The purpose is to advance the development of restoration techniques that are suitable for the practical short-form UGC platforms. Thus, we cooperated with the  Kwai Platform and collected our KwaiSR dataset, which is composed of 1,800 synthetic image pairs and 1,900 low-quality wild images. We divided the KwaiSR into training, validation, and testing sets for our challenge, respectively. The synthetic dataset simulates real-world degradation patterns, while the wild dataset includes low-quality images filtered using the KVQ quality assessment method. We organized the NTIRE 2025 challenge based on the collected datasets, intending to attract researchers to address this challenging problem. The challenge results highlight a promising direction for advancing image super-resolution.


\section*{Acknowledgments}
This work was partially supported by NSFC under Grant 623B2098 and the China Postdoctoral Science Foundation-Anhui Joint Support Program under Grant Number 2024T017AH. We thank Kuaishou for providing the source images of this dataset and sponsoring this challenge. 
{
    \small
\bibliographystyle{ieeenat_fullname}
    \bibliography{main}
}


\end{document}